\pdfoutput=1
\documentclass[11pt]{article}

\usepackage[final]{acl}

\usepackage{times}
\usepackage{latexsym}

\usepackage[T1]{fontenc}

\usepackage[utf8]{inputenc}
\usepackage{amsmath}
\usepackage{multirow}
\usepackage{multicol}
\usepackage{amssymb}
\usepackage{morewrites}
\usepackage{etex}
\usepackage{dashrule}
\usepackage{booktabs}
\usepackage{arydshln}

\usepackage{microtype}
\usepackage{mdframed}
\usepackage{fancyvrb}
\usepackage{inconsolata}

\usepackage{graphicx}
\usepackage{enumitem}
\setlist[itemize]{nosep,leftmargin=*}
\DefineVerbatimEnvironment{myverbatim}{Verbatim}{breaklines=true}

\title{\textsc{BPO}: Towards Balanced Preference Optimization between Knowledge Breadth and Depth in Alignment}

\author{
    \textbf{Sizhe Wang\textsuperscript{1} \quad Yongqi Tong\textsuperscript{2} \quad Hengyuan Zhang\textsuperscript{3}} \\
    \textbf{\quad Dawei Li\textsuperscript{4} \quad Xin Zhang\textsuperscript{2} \quad Tianlong Chen\textsuperscript{5}} 
    \\
    \textsuperscript{1}University of Southern California, 
    \textsuperscript{2}Ant Group,
    \textsuperscript{3}Tsinghua University \\
    \textsuperscript{4}Arizona State Univeristy,  
    \textsuperscript{5}University of North Carolina at Chapel Hill
    \\
    \texttt{sizhewan@usc.edu}, \texttt{\{tongyongqi.yq, evan.zx\}@antgroup.com},\\
    \texttt{zhang-hy22@mails.tsinghua.edu.cn}, \\ \texttt{daweili5@asu.edu}, \texttt{tianlong@cs.unc.edu}
}

\begin{document}
\maketitle
\begin{abstract}
Reinforcement Learning with Human Feedback (RLHF) is the key to the success of large language models (LLMs) in recent years. 
In this work, we first introduce the concepts of knowledge breadth and knowledge depth, which measure the comprehensiveness and depth of an LLM or knowledge source, respectively.
We reveal that the imbalance in the number of instructions and responses can lead to a potential disparity in breadth and depth learning within alignment tuning datasets by showing that even a simple uniform method for balancing the number of instructions and responses can lead to significant improvements. Building on this, we further propose \underline{\textbf{B}}alanced \underline{\textbf{P}}reference \underline{\textbf{O}}ptimization (\textbf{\textsc{BPO}}), designed to dynamically augment the knowledge depth of each sample. \textsc{BPO} is motivated by the observation that the usefulness of knowledge varies across samples, necessitating tailored learning of knowledge depth. 
To achieve this, we introduce gradient-based clustering, estimating the knowledge informativeness and usefulness of each augmented sample based on the model's optimization direction. Our experimental results on various benchmarks demonstrate that \textsc{BPO} outperforms other baseline methods in alignment tuning while maintaining training efficiency. Furthermore, we conduct a detailed analysis of each component of \textsc{BPO}, providing guidelines for future research in preference data optimization.
Our code will be published soon on \texttt{\url{https://github.com/Sizhe04/Balanced-Preference-Optimization}}.


\end{abstract}

\section{Introduction}
\begin{figure}[h]
    \centering
    \scalebox{0.28}{\includegraphics{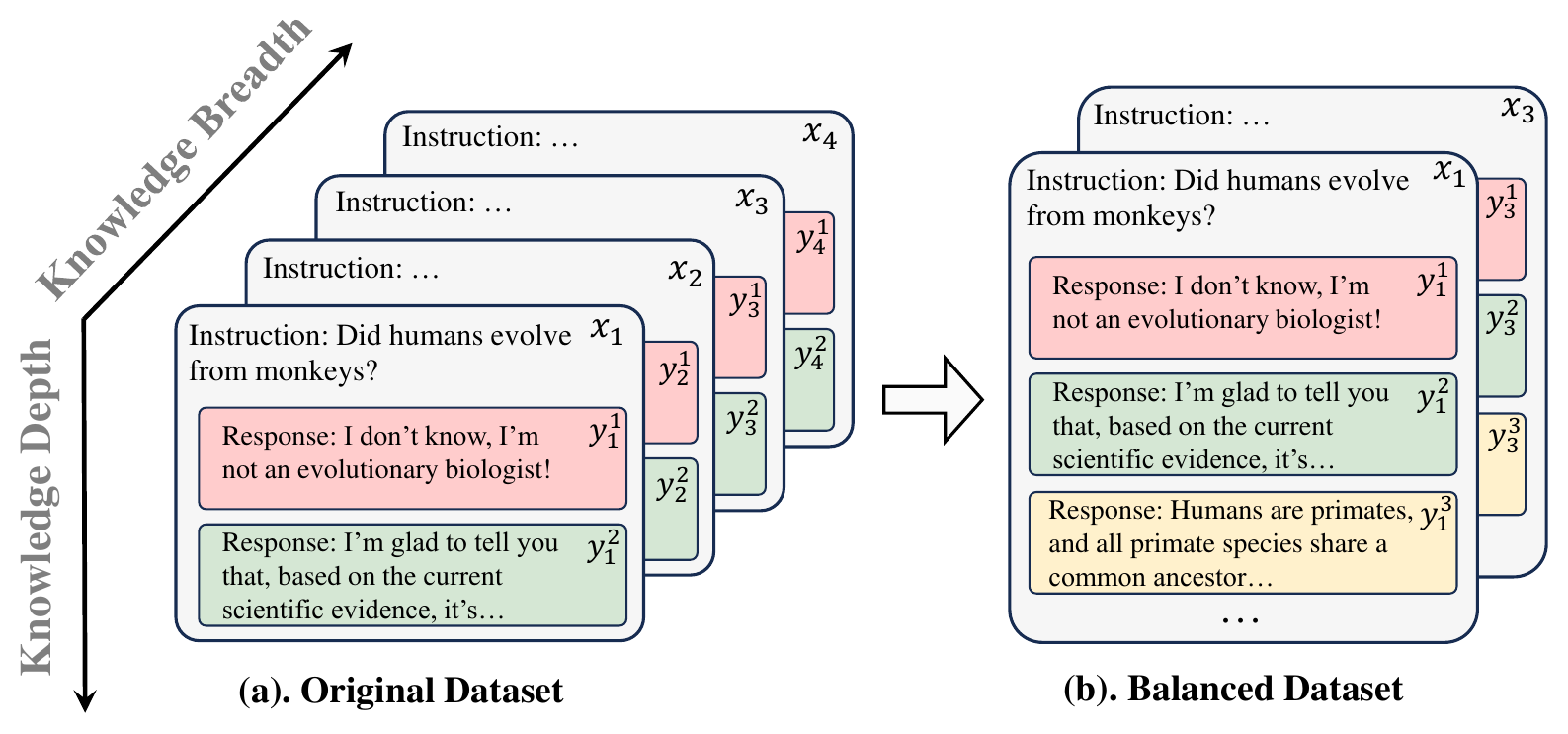}}
    \caption{Overview of knowledge breadth and depth, and how we link them with the number of instructions and responses in alignment tuning datasets.}
    \label{fig:learng space}
\end{figure}


Reinforcement Learning with Human Feedback (RLHF)~\cite{christiano2017deep} has played a pivotal role in the success of large language models (LLMs) in recent years.
It aims to align LLMs with human values and preferences during the post-training phase, leveraging extensive pairwise feedback from human annotators.
To boost this alignment tuning, various advanced training techniques have been introduced, including Direct Preference Optimization (DPO)~\cite{rafailov2024direct}, Sequence Likelihood Calibration (SLiC)~\cite{zhao2022calibrating}, and Kahneman-Tversky Optimization (KTO)~\cite{ethayarajh2024kto}.


Another line of effort aims to enhance the alignment process from a data perspective.
This involves techniques such as data selection~\cite{zhou2024lima}, sampling~\cite{khaki2024rs}, and rearrangement~\cite{pattnaik2024curry} to enhance various aspects of the original preference learning dataset, including quality~\cite{zhou2024lima}, diversity~\cite{song2024scaling}, and difficulty~\cite{liu2023makes,pattnaik2024curry}.
While existing data optimization methods focus on individual components of preference learning data (either the instruction or the response), a unified approach to systematically optimize preference data as a whole remains lacking.



To address this problem, we begin by proposing the concepts of \textit{knowledge breadth} and \textit{knowledge depth}, which measure the comprehensiveness and depth of an LLM or knowledge source respectively.
Building on these concepts, we conduct a systematic analysis of the resources allocated for enhancing the breadth and depth of knowledge during alignment tuning accordingly, uncovering a potential imbalance stemming from the structure of alignment datasets.
To validate this imbalance, we propose a simple balance method, that includes knowledge breadth compression (KBC) and knowledge depth augmentation (KDA), achieved by uniformly reducing the number of prompts and increasing the number of response pairs in the preference learning dataset.
Figure~\ref{fig:learng space} demonstrates the overview process.
The substantial improvement brought by both KBC and KDA confirms the effectiveness of our balance approach in alignment tuning datasets.



We further examine the preference learning dataset and observe the need to dynamically augment the knowledge depth of different samples. Some samples in preference data contain rich, informative knowledge, while others consist of relatively shallow or less useful information. The former deserves more response pairs to allow LLMs to fully learn from them, whereas prior studies~\cite{zhou2024lima} have shown that excessive focus on less useful knowledge can lead to negative outcomes. 
Therefore, we propose \textbf{B}alanced \textbf{P}reference \textbf{O}ptimization (\textbf{\textsc{BPO}}), a method designed to dynamically augment the knowledge depth of each sample based on its informativeness.
\textsc{BPO} first follows a general pipeline~\cite{albalak2024survey,xia2024less} to sample a subset of representative and diverse prompts by clustering.
Next, we introduce gradient-based~\cite{pruthi2020estimating,han2023understanding} clustering to estimate the optimal knowledge depth for each sample from the perspective of model optimization.
By clustering each augmented \textit{(prompt-response pair)} based on their gradient features and selecting those closest to the cluster centers, \textsc{BPO} ensures that samples with gradient features near the centroids receive more focus. These centrally positioned samples are typically more informative and helpful for model optimization, making them deserving of greater knowledge-depth learning resources.


Extensive experiments across various benchmarks demonstrate both the effectiveness and computational efficiency of our \textsc{BPO} approach compared to other baseline methods. Additional ablation studies and hyperparameter analyses confirm the effectiveness of each component, highlighting \textsc{BPO}’s robustness across different hyperparameter settings. We also investigate alternative methods for dynamically estimating the required knowledge depth, with experimental results showing that the gradient-based approach performs best, achieving the highest human preference scores.


To summarize, our contribution in this work is threefold:

\begin{itemize}
    \item We introduce the concept of balancing knowledge breadth and depth, and conduct preliminary experiments to validate the effectiveness of this balancing in alignment tuning datasets.
    \item We propose \textsc{BPO}, which involves estimating samples' difficulties and dynamically augmenting the knowledge depth for each sample via hierarchical sampling.
    \item Through extensive experiments, we show that \textsc{BPO} outperforms previous preference data optimization methods. We also conduct further exploration on \textsc{BPO} to provide in-depth hints.
\end{itemize}


\begin{figure*}[!th]
    \centering
    \centerline{\includegraphics[width=2\columnwidth]{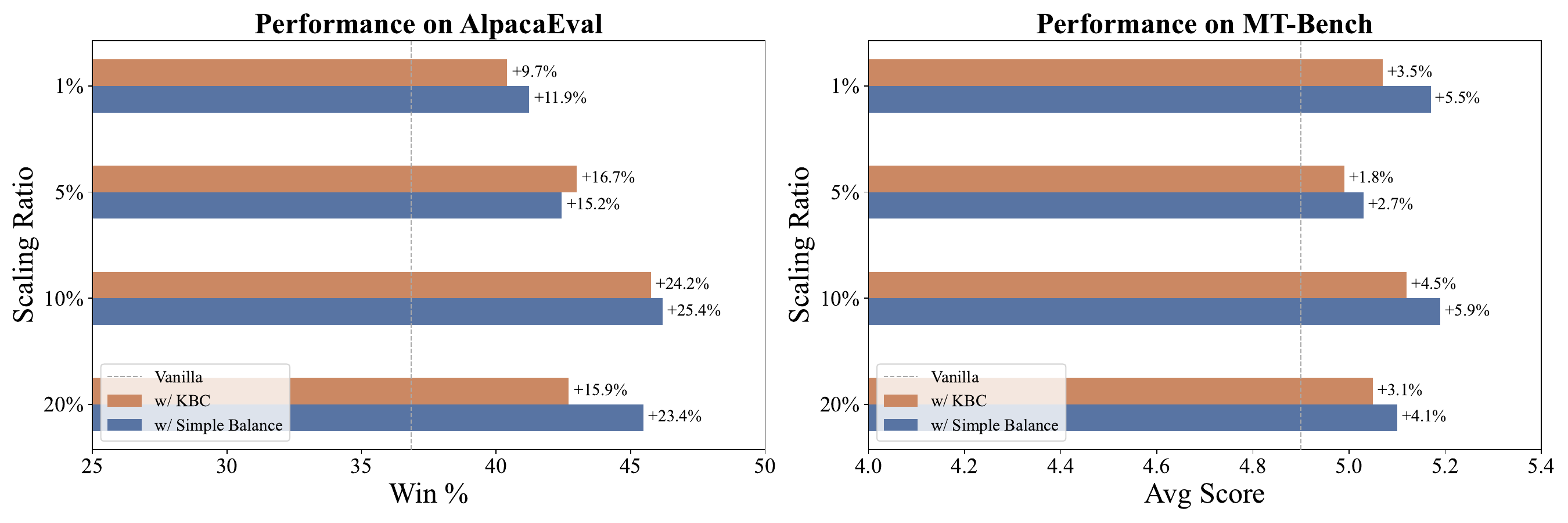}}
    \caption{
    Preliminary experiment results on SafeRLHF using simple balance.
    }
    \label{fig:preliminary}
\end{figure*}


\section{Preliminary}
\label{preliminary}

In this section, we first provide the definition of knowledge breadth and depth and bridge the connection between the two concepts with instruction and response numbers in alignment tuning datasets. Then, we propose a simple strategy to balance the knowledge breadth and depth in the DPO tuning process and conduct preliminary experiments to evaluate its effectiveness.

\subsection{Knowledge Breadth and Depth}
In general, knowledge breadth \( \mathcal{B} \) represents the model's range of knowledge across various subjects or domains, which is a measure of how many different areas or topics the model can understand and provide information on. In contrast, knowledge depth \( \mathcal{K} \) refers to how well an LLM can provide in-depth and detailed information on specific topics. It is a measure of the model’s ability to delve into complexities, offer nuanced insights, and demonstrate expertise in narrow subject areas~\cite{bai2024benchmarking}.
Thus, we can simply define a knowledge source \( \mathcal{L} \) as \( \mathcal{L} = (\mathcal{B}, \mathcal{K}) \).

While LLMs demonstrate striking knowledge breadth in extensive domains and areas, their knowledge depth for providing truly in-depth and expert-level output is still not promising~\cite{zhang2024balancing,bai2024benchmarking}.
In this work, we propose to address this problem by first analyzing the relationship between knowledge breadth and depth, with prompts and responses in the alignment tuning dataset.
Intuitively, a large and diverse set of instructions allows the LLMs to cover a wide range of topics and areas, thus expanding its knowledge breadth. 
Similarly, more responses of various quality provide LLMs with great opportunities to fully understand the question and explore the required knowledge, thus leading to LLMs being able to provide in-depth insights and responses.

However, if we take a closer look at alignment tuning datasets, there is a significant imbalance between the number of instructions and responses: a typical alignment tuning dataset is in the format of 
\( \mathcal{D} = \{(x_{1}, y_{1}^{1}, y_{1}^{2}), ..., (x_{n}, y_{n}^{1}, y_{n}^{2})\} \), in which each sample (\(x, y^{1}, y^{2}\)) encompasses a prompt \(x\), the winning response \(y^{1}\) and the losing response \(y^{2}\). In this dataset, the instruction number $n$ is usually in the tens of thousands, while the response number is simply 2, where we have $n>>2$. Here we argue that this imbalance between instruction and response numbers actually implies the unbalanced resource allocation for knowledge breadth and depth learning during alignment tuning time, limiting LLMs' exploration for more in-depth knowledge.

\subsection{Simple Balance}
To validate our hypothesis in knowledge breadth and depth and further explore the impact of instruction and response number in LLM alignment tuning, we propose a simple balance method.
This method involves first conducting knowledge breadth compression (KBC) by clustering and sampling a subset of instruction.
After that, knowledge depth augmentation (KDA) is adopted to produce synthetic response pairs to uniformly augment knowledge depth.
We provide more details in Section~\ref{Knowledge Breadth Compression} and Section~\ref{Dynamic Knowledge Depth Augmentation}.
Thus, we uniformly balance knowledge breadth and depth from 
\( \mathcal{L} = (n, 2) \) to \( \mathcal{L}_{bal} = (n \times s, \frac{2}{s}) \), where \( s \in (0, 1] \) is the scaling ratio for balancing.



\begin{figure*}[h]
    \centering
    \scalebox{0.32}{\includegraphics{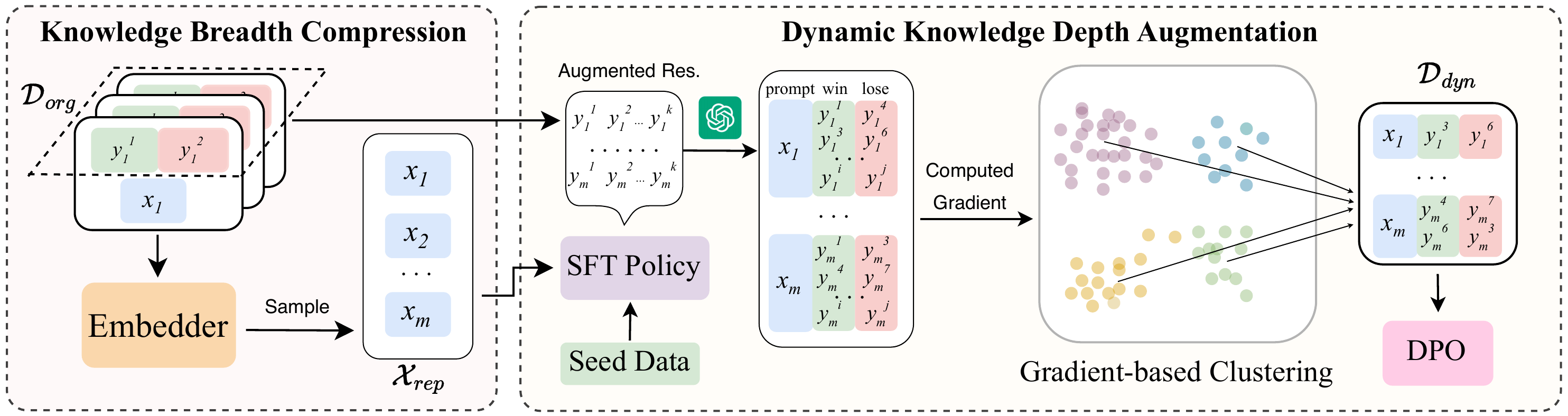}}
    \caption{Overview of \textsc{BPO} pipeline. \textsc{BPO} first selects representative prompts to reduce knowledge breadth through embedding-based clustering. Next, it generates responses using the SFT policy and employs GPT-4 to score these responses to uniformly construct response pairs. Subsequently, \textsc{BPO} samples response pairs to dynamically augment knowledge depth through gradient-based clustering. Finally, DPO is applied to the sampled data, which ensures efficient alignment.}
    \label{fig:BPO}
\end{figure*}

\subsection{Preliminary Experiment \& Analysis}
Preliminary experiment aims to investigate the results of compressing knowledge breadth \( \mathcal{B} \) with various scaling ratios while correspondingly and uniformly increasing knowledge depth \( \mathcal{K} \).
We follow the experimental and evaluation settings in Section~\ref{experimental setup}.

In Figure~\ref{fig:preliminary}, we demonstrate that using embedding-based clustering to select only 1\% to 10\% of the representative instructions for alignment can achieve performance comparable to, and even surpassing, alignment on the full dataset while significantly reducing computational resource requirements.
This finding is consistent with that of~\citet{zhou2024lima}, indicating that data quantity is not the key factor. 
By focusing on samples at the centroids of clusters, we ensure the quality of the selected prompts while reducing the probability of redundant prompts being selected, thus achieving better performance with fewer prompts.
This indicates that a small subset of carefully selected instructions could already satisfy the need of LLMs to expand the breadth of knowledge.

Furthermore, augmenting the knowledge depth of LLMs by constructing diverse responses for these selected prompts leads to further improvements in model performance. 
We also conduct experiments on HHRLHF dataset, as shown in Appendix~\ref{app:preliminary}. 
These results highlight that balancing knowledge breadth and depth by dataset manipulation enables a more efficient and effective alignment of the model.

\section{Our Method: Balanced Preference Optimization}
\label{our method}
In this section, we introduce our \textsc{BPO} method, which involves two core steps: Knowledge Breadth Compression (Section~\ref{Knowledge Breadth Compression}) and Dynamic Knowledge Depth Augmentation (Section~\ref{Dynamic Knowledge Depth Augmentation}).

\subsection{Knowledge Breadth Compression}
\label{Knowledge Breadth Compression}

To reduce the breadth \( \mathcal{B} \) of the knowledge, we aim to select the most representative subset of prompts \( \mathcal{X_\text{rep}} \subset \mathcal{D}_\text{org} \).
We follow the previous works and adopt a clustering method based on the prompts' embedding vectors.
Specifically, we first obtain the embedding vector $e_i$ using a pre-trained embedded $f$ and have $e_i = f(x_i)$.

We then apply the \textit{K}-means clustering algorithm to the set of embedding vectors \( \{e_1, e_2, ..., e_n\} \) to partition them into 
\(C\) clusters \( \{ \mathcal{C}_1, \mathcal{C}_2, ..., \mathcal{C}_C \} \). 
Within each cluster \( \mathcal{C}_c \), we select the top \( s \) fraction of prompts, where \(s\) is the scaling ratio,
based on Euclidean distance \( d(e_i, \mu_c) \), whose embedding vectors are closest to the centroid \(\mu_c\) and intuitively are the most representative:
\begin{equation} 
\small
\mathcal{X}_{rep} = \bigcup_{c=1}^{C} \left\{ x_i \in \mathcal{C}_c \ \bigg| \ \text{rank}\left( d(\mathbf{e}_i, \boldsymbol{\mu}_c) \right) \leq s \cdot |\mathcal{C}_c| \right\}
\label{rank}
\end{equation}


\subsection{Dynamic Knowledge Depth Augmentation}
\label{Dynamic Knowledge Depth Augmentation}

\paragraph{Knowledge Depth Augmentation}

To augment the knowledge depth \( \mathcal{K} \) for a given preference learning dataset, we first tune the base model with LoRA~\cite{hu2021lora} using a subset of seed data and obtian the supervised fine-tuned (SFT) policy \( \pi_\theta \), more details are in Appendix~\ref{preSFT}. Then, we produce multiple responses for each prompt \(x_i \in \mathcal{X}_{\text{rep}}\) using the SFT policy \( \pi_\theta \) and follow previous work to adopt the LLM-as-a-judge~\cite{zheng2023judging,gao2023human,li2024dalk} and reject sampling~\cite{khaki2024rsdpohybridrejectionsampling} approach to construct high-quality response pairs.
Furthermore, since the SFT policy has already been optimized for common scenarios, its responses are typically safe and exhibit minimal diversity, making it challenging to generate response pairs with significant score differences. 
To address this for the safety dataset, we employed a novel method by utilizing jailbreaking prompts to elicit more diverse responses. 
We analyze this method and its effects in Section~\ref{comparision online generation}.

\paragraph{Gradient Computation and Projection}
 
Building upon the previously established uniform depth augmentation, we dynamically allocate depth for each prompt \(x_i\) by utilizing the gradient features of each \textit{(prompt-response pair)}. Following the approach outlined by \citet{xia2024less}, we employ Low-Rank Adaptation (LoRA) during the initial supervised fine-tuning (SFT) step to minimize the number of trainable parameters. Additionally, we implement random projection \citep{johnson1984extensions} to reduce the dimensionality of the LoRA gradients. Specifically, the parameter update process, based on the Adam optimizer \citep{kingma2014adam}, is as follows:
\begin{equation}
\textstyle
\theta^{t+1} - \theta^t = -\eta_t \Gamma(z, \theta^t)
\end{equation}
\begin{equation}
\textstyle
\Gamma(z, \theta^t) \triangleq \frac{m^{t+1}}{\sqrt{v^{t+1}} + \epsilon}
\end{equation}
\begin{equation}
\textstyle
m_{t+1} = \beta_1 m_t + (1 - \beta_1) \nabla \ell(z, \theta_t)
\end{equation}
\begin{equation}
\textstyle
v_{t+1} = \beta_2 v_t + (1 - \beta_2) (\nabla \ell(z, \theta_t))^2
\end{equation}
where \(\beta_1, \beta_2\) are the hyperparameters, \(\epsilon\) is a small constant, and \(\Gamma(z, \theta^t)\) represents the first-order expansion for the Adam dynamics.
Subsequently, for a given data sample \(z\) and model checkpoint \( \theta^t \), we use random projection \citep{johnson1984extensions} to map the high-dimensional LoRA gradients \(\nabla \ell(z; \theta^t)\) into a lower \textit{d}-dimensional space, denoted as \( \widehat{\nabla} \ell (z; \theta^t) \).
This accelerates the following clustering process and more details are provided in Appendix \ref{app:random project}.
This gradient-based data representation allows us to measure the contribution of each response pair to the model's optimization, enabling more efficient learning by dynamically adjusting the allocated knowledge depth resource for each prompt.

\paragraph{Dynamic Depth Allocation}

\begin{figure}[h]
    \centering
    \scalebox{0.25}{\includegraphics{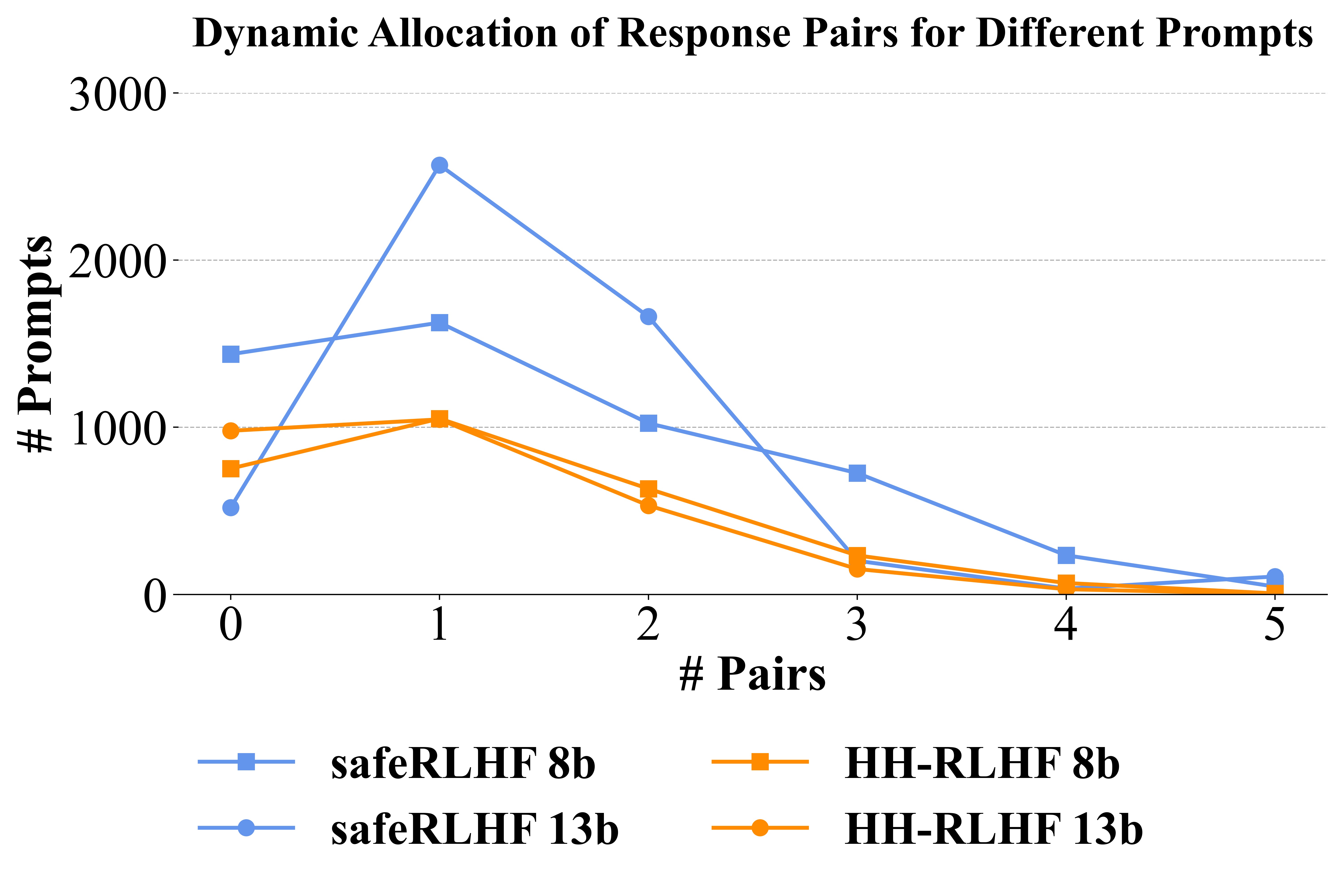} }
    \caption{
Dynamic allocation of response pairs based on gradient clustering. 
Different prompts require varying numbers of pairs. 
While most prompts can be adequately represented with a single response pair, certain prompts demand a more comprehensive exploration involving additional pairs.
    }
    \label{fig:dynamic}
\end{figure}

We perform \textit{K}-means clustering on the set of gradient features \( \widehat{\nabla} \ell (z; \theta_t) \) into \( G \) clusters \( \{ G_1, G_2, \ldots, G_G \} \) to evaluate the required knowledge depth for each sample from the model optimization perspective. 
For each cluster \( G_g \), we select the top \( \eta \% \) of \textit{(prompt, response pair)} whose gradient vectors are closest to the cluster centroid \( \boldsymbol{\mu}_g \), based on the Euclidean distance:
\begin{equation} 
\small
\begin{split}
\mathcal{D}_{dyn} = \bigcup_{g=1}^{G} \left\{ z_i \in \mathcal{G}_g \ \bigg| \ \text{rank}\left( d(\widehat{\nabla} \ell (z; \theta^t), \boldsymbol{\mu}_g) \right) \leq \eta \cdot |\mathcal{G}_g| \right\}
\end{split}
\end{equation}
More results and analysis about the hyperparameter \( \eta \% \) are shown in Table~\ref{ab:clustering about kpa}. 
Intuitively, informative augmented samples with gradient vectors close to the centers will get a greater opportunity to be selected, thus being allocated with more depth learning resources.
Figure~\ref{fig:dynamic} shows the dynamic allocation of response pairs for different prompts, where the number of pairs varies based on the gradient-based clustering and selection. 
This illustrates how \textsc{BPO} dynamically adjusts the number of response pairs for different samples to optimize the knowledge depth allocation.
We then perform DPO fine-tuning using $\mathcal{D}_{dyn}$, leading to better alignment and performance improvements as demonstrated in our experiments.

\begin{table*}[h]
\centering
\small
\begin{tabular}{ccccccccc}
\toprule[2pt]
\multirow{2}{*}{\textbf{Model}} & \multirow{2}{*}{\textbf{Method}} & \multicolumn{3}{c}{\textbf{safeRLHF}} & & \multicolumn{3}{c}{\textbf{HH-RLHF}} \\ \addlinespace[2pt] \cline{3-5}  \cline{7-9} \addlinespace[2pt]
 &  & \textbf{Data Size} & \textbf{MT-Bench}  & \textbf{AlpacaEval} & & \textbf{Data Size} & \textbf{MT-Bench}  & \textbf{AlpacaEval} \\ 
 \addlinespace[2pt] \hline  \addlinespace[2pt]
 
\multirow{7}{*}{$\text{Llama-3}_{\text{8B}}$}      
& Vanilla DPO     & 50,489     & 4.90   & 36.84\%   &   & 27,000          & 4.71   & 18.14\%     \\ 
& KBC (\(s\)=10\%)  & 5,094     & 5.12   & 45.75\%    &   & 2,744          & 4.11   & 20.12\%     \\ 
& Simple Balance & 45,522     & 5.19   & 46.18\%   &   & 32,928          & 4.42   & 25.34\%     \\ 
& Simple Balance [1]    & 4,552     & 5.02   & 43.48\%    &   & 3,292          & 4.21   & 23.75\%     \\ 
& Curry-DPO            & 45,522    & 5.21   & 44.03\%    &   & 32,928   & 4.67   & 28.29\%     \\
& RS-DPO               & 52,067    & 5.32   & 47.48\%    &   & 34,177       & 5.12   & 33.31\%     \\ 
& \textsc{BPO} (Ours)  & 4,570      & \textbf{5.45}   & \textbf{48.24\% } &  & 3,312     & \textbf{5.48 }  & \textbf{40.25\%}
\\ \addlinespace[2pt] \hdashline[1pt/1pt] \addlinespace[2pt]

\multirow{7}{*}{$\text{Llama-2}_{\text{13B}}$} 
& Vanilla DPO         & 50,489     & 5.40   & 48.07\%  &  & 27,000        & 5.79   & 21.50\%     \\ 
& KBC (\(s\)=10\%)     & 5,094      & 5.38   & 46.71\%  &  & 2,744          & 5.14   & 22.85\%     \\ 
& Simple Balance      & 29,555     & 5.68   & 50.30\%  &  & 26,792        & 5.51   & 24.68\%     \\ 
& Simple Balance [1]        & 2,955      & 4.02   & 48.64\%  &  & 2,679         & 4.64   & 21.11\%     \\ 
& Curry-DPO            & 29,555     & 5.67   & 48.50\%  &  & 26,792        & 5.59   & 24.22\%     \\
& RS-DPO               & 49,664     & 5.79   & 50.73\%  &  & 27,143        & \textbf{5.83}   & 25.26\%     \\ 
& \textsc{BPO} (Ours)  & 3,042     & \textbf{5.85}   & \textbf{54.27\%}  & & 2,753          & 5.74   & \textbf{26.96\%}
\\ \bottomrule[2pt]
\end{tabular}
\caption{Performance comparison of different pair selection strategies on the SafeRLHF and HH-RLHF datasets. 
[1] stands for utilizing Simple Balance on 10\% randomly selected samples.
As shown, our method \textsc{BPO} can achieve comparable or even better performance with no more than 10\% overall data. 
}
\label{tab: main exp}
\end{table*}

\section{Experiments}
In this work, we introduce our experiment settings and analyze our results. 

\subsection{Experimental Setup}
\label{experimental setup}
\paragraph{Datasets}
We choose \textbf{HH-RLHF}~\cite{bai2022training} and \textbf{safeRLHF}~\cite{dai2023safe} as our training sets.
HH-RLHF consists of two subsets, with approximately 170K chosen-rejected pairs related to human and AI assistant dialogues. For our experiment, we randomly select 30,000 samples from the helpfulness subset. 
SafeRLHF contains 83.4K preference entries, where each entry includes two responses with labels for harmlessness and helpfulness. 
In our experiment, we select samples where both labels are the same, resulting in approximately 55,000 samples.
Additionally, for each dataset, we randomly select 10\% of the dataset \( \mathcal{D}_{org} \) as the seed data \( \mathcal{D}_{seed} \), using annotated chosen responses as golden reference to perform preliminary SFT.



\paragraph{Baselines} Two recent preference data optimization methods are used to compare with ours in this domain: \textbf{RS-DPO}~\citep{khaki2024rsdpohybridrejectionsampling} and \textbf{Curry-DPO}~\citep{Pattnaik2024CurryDPOEA}. 
RS-DPO initially generates diverse responses from a supervised fine-tuned model. Then it guides a reward model to identify contrastive pairs based on its reward distribution, and applies DPO to optimize the model's alignment. In our experiment, we set the threshold of scoring variances as 0.9.
Curry-DPO leverages multiple preference pairs per prompt, structured through curriculum learning from easy to hard tasks. 
It systematically ranks these pairs based on response quality differences and iteratively trains the model, resulting in improved performance across various benchmarks compared to standard DPO.



\paragraph{Implementation Details} 

We conduct experiments using Llama-2-13B~\citep{touvron2023llama} and Llama-3-8B~\citep{dubey2024llama}. 
We also utilize Llama3-8b as the feature extractor to obtain representations used for clustering, which serves as a common choice in many clustering work~\citep{petukhova2024text}. 
We adopt the \textbf{MT-Bench}~\citep{zheng2023judging} and \textbf{AlpacaEval}~\citep{li2023alpacaeval} to evaluate the effectiveness of our proposed methods and baseline models.
We set the scaling ratio for knowledge breadth and depth balance to 0.1.
More evaluation details are attached in Appendix~\ref{dataset} and the training hyperparameters are shown in Appendix~\ref{hyper}. 
Additionally, we conducted further experiments on other datasets to validate the effectiveness and robustness of our method, which are detailed in Appendix~\ref{additional_exp}.

\begin{figure*}[!th]
    \centering
    \centerline{\includegraphics[width=2\columnwidth]{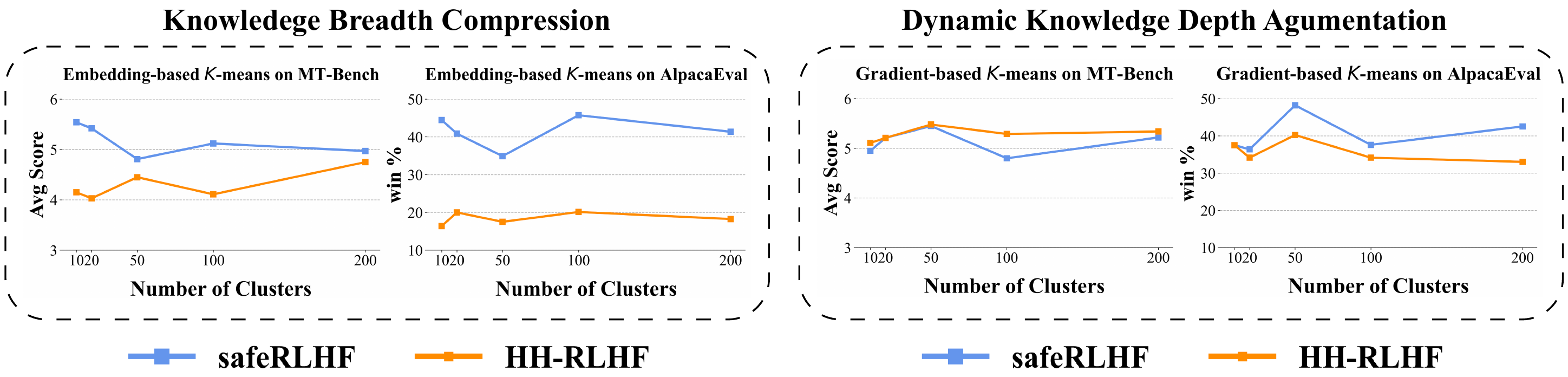}}
    \caption{
    Experimental results for varying numbers of clusters during embedding-based and gradient-based \textit{K}-means
    The top \( \eta = 10\% \) of data points are selected based on Equation \ref{rank} for all clustering tasks.
    All experiments are conducted on $\text{Llama-3}_\text{8B}$.
    }
    \label{fig:ab_num_clusters}
\end{figure*}

\subsection{Experimental Results}

In this section, we compare our overall \textsc{BPO} methods with the mentioned baselines and sampling methods. 
As shown in Table~\ref{tab: main exp}, \textsc{BPO} BPO achieves comparable or even superior performance while using no more than 10\% of the data.
KBC underperforms compared to \textsc{BPO} as it only relies on using fewer representative instructions, which further demonstrate the necessity for our hierichical sampling approach to balance between both breadth and depth. 
Compared to simple balance, BPO achieves better performance, which we attribute to the gradient-based dynamic knowledge depth augmentation selecting the most useful and informative samples for model updates.
Additionally, potentially redundant samples are less likely to be selected during sampling. This reduces the possible negative impact of knowledge on the model and increases training efficiency.
Furthermore, we compared our method with the approach of randomly selecting 10\% of the samples used in simple balance.
The results show that our gradient clustering performs better, indicating that BPO successfully allocates appropriate knowledge depth resources, thereby enhancing model performance.
Furthermore, we compared our method with Curry-DPO and RS-DPO. Notably, RS-DPO's performance ranks second only to ours, even surpassing BPO on MT-Bench when trained on the HH-RLHF dataset. However, it requires significantly more data than our approach. Overall, our method is both effective and efficient, demonstrating robustness across two datasets and two model sizes while successfully achieving alignment.


\begin{figure*}[!th]
    \centering
    \centerline{\includegraphics[width=2\columnwidth]{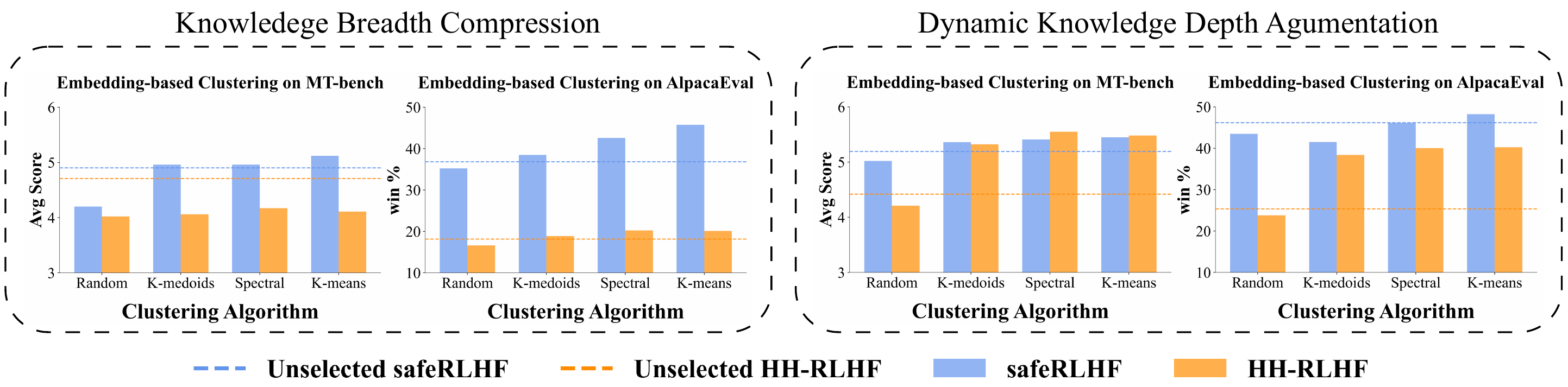}}
    \caption{
    Experimental results for different clustering algorithms and random selection used in knowledge breadth compression and dynamic knowledge depth augmentation.
    For knowledge breadth compression (left part), the number of clusters is set to 100, and for dynamic knowledge depth augmentation (right part), it is set to 50.
    The top \( \eta = 10\% \) of data points are selected based on Equation \ref{rank} for all clustering, maintaining the same percentage for random selection.
    All experiments are conducted on $\text{Llama-3}_\text{8B}$.
    }
    \label{fig:ab_cluster_method}
\end{figure*}

\section{Extra Investigation \& Further Analysis}
\label{sec:extra}

In this section, we explore how different clustering configurations affect \textsc{BPO}'s performance, compare approaches for measuring knowledge depth, and analyze various response generation methods.

\subsection{Clustering Settings}
\paragraph{Number of Clusters}
We investigate how the number of clusters affects both embedding vector clustering during the knowledge breadth compression step and gradient clustering in the dynamic knowledge depth augmentation step.
Figure~\ref{fig:ab_num_clusters} shows that when using 100 clusters for embedding vector clustering and 50 clusters for gradient clustering, the results are relatively stable.
A cluster number that is too small may fail to adequately represent the comprehensive level of knowledge breadth, making it difficult to achieve stable results and leading to significant fluctuations.
Conversely, an excessively large number of clusters might over-segment the data, which does not necessarily guarantee a steady improvement in performance.

\paragraph{Clustering Algorithm}
We explore the influence of different clustering algorithms on the performance of \textsc{BPO}.
In addition to \textit{K}-means, we apply \textit{K}-medoids and spectral clustering.
Figure~\ref{fig:ab_cluster_method} demonstrates that the performance difference between \textit{K}-means and spectral clustering is minimal, while \textit{K}-medoids performs slightly worse than the others.
Furthermore, these clustering-based selection methods consistently outperform random selection, and except in rare cases, generally achieve better performance than using all data without selection.
This further highlights the robustness of our BPO in various clustering algorithm selections.

\subsection{Measurement of Required Knowledge Depth}
In our \textsc{BPO}, we use a gradient-based method to estimate the required depth of knowledge for each sample. 
To ensure its efficacy and our method's generalization, we also explore additional measurement approaches from various perspectives, including length and semantic similarity.

\paragraph{Length}
To evaluate required knowledge depth, we measure the length of generated responses. 
We assume that high variability in response lengths for an instruction potentially may indicate inconsistent understanding, suggesting a need for deeper learning. 
Consequently, prompts with greater variance in response length are assigned greater depth. 
Further details are provided in Appendix \ref{ab:length}.

\paragraph{Semantic Similarity}
We also utilize semantic similarity to measure required knowledge depth \citep{petukhova2024text}. 
We assume that high similarity in responses intuitively indicates a stable understanding, requiring less learning.
Conversely, low similarity suggests inconsistent understanding, necessitating deeper learning.
Further details are provided in Appendix \ref{ab:semantic}.

\paragraph{Measurement Approach Comparisons} 
Table \ref{ab:measure} illustrates that relying solely on response length is inadequate, as it overlooks semantic differences. 
Responses of similar length can have different meanings, leading to an incomplete exploration of the data. 
Although semantic similarity performs better than response length, it still underperforms our gradient-based methods. 
While semantic similarity captures the correlations within the responses of each prompt, gradient feature represents knowledge breadth and depth on a global scale, resulting in more effective optimization.

\begin{table}[!th]
\tiny
\setlength\tabcolsep{1.5pt}
\fontsize{7.5}{9}\selectfont 
\centering
\begin{tabular}{lccccc}
\toprule[2pt]
\multirow{2}{*}{\textbf{Measurement}} & \multicolumn{2}{c}{\textbf{SafeRLHF}} & & \multicolumn{2}{c}{\textbf{HH-RLHF}} \\ 
\cline{2-3} \cline{5-6} \addlinespace[2pt]
& \textbf{MT-Bench} & \textbf{AlpacaEval} & & \textbf{MT-Bench} & \textbf{AlpacaEval} \\  \addlinespace[2pt] \hline \addlinespace[2pt] 
Length & 4.48 & 34.85\%  & & 4.27 & 24.61\% \\ 
 Semantic Similarity & 4.92 & 40.87\% & & 4.68 & 29.75\% \\ 
 Gradient & 5.45 & 48.24\% & & 5.48 & 40.25\% \\ \bottomrule[2pt]
\end{tabular}
\caption{Performance comparison of different methods for measuring the required knowledge depth. 
The results demonstrate that measuring via gradients in our \textsc{BPO} leads to better performance. 
}
\label{ab:measure}
\end{table}

\subsection{Comparisons about Augmentation Methods in KDA}
\label{comparision online generation}

As described in Table~\ref{ab:safety}, common response generation methods adopted in many online DPO variants on the safeRLHF dataset surprisingly did not yield promising results. 
Upon analyzing the bad cases, we found that the base model had already been optimized for common scenarios, making its responses quite safe and difficult to obtain unsafe ones. 

Therefore, we employed a different method in the safety alignment area: for half of the generations, we used jailbreaking prompts proposed in~\citet{shen2024donowcharacterizingevaluating} to induce the model to produce potentially risky responses. 
These were then paired with normally generated safe responses for further processing. 
The pair construction process is the same as described in Section~\ref{Dynamic Knowledge Depth Augmentation}.
Our findings also support the assumption that uniformly increasing the number of more distinctly different pairs between chosen and rejected responses can notably enhance the model alignment effects.  
Moreover, in the context of online RLHF research focused on safety alignment, we want to emphasize to future research the importance of assessing quality differences in generations. 
It is crucial to determine whether the model is already sufficiently safe, as this could hinder the generation of adequately poor negative responses.

\begin{table}[ht]
\small
\centering
\begin{tabular}{ccc}
\toprule[2pt]
\textbf{Generation Methods} & \textbf{MT-Bench} & \textbf{AlpacaEval} \\  \addlinespace[2pt] 
 \hline \addlinespace[2pt] 
w/o data generation & 5.12 & 45.75\% \\ 
normal generations & 4.40 & 31.18\% \\ 
half normal / half jailbreak & 5.19 & 46.18\% \\ 
 \bottomrule[2pt]
\end{tabular}
\caption{
$\text{Llama-3}_\text{8B}$'s performance of different online generation methods in Dynamic Knowledge Depth Augmentation step on the safeRLHF dataset. 
}
\label{ab:safety}
\end{table}

\section{Related Work}
\label{relatedwork}

\textbf{Preference Learning.}
Preference Learning aims to align LLMs' output distribution with human preferences and values. 
The most widely used alignment method is Reinforcement Learning from Human Feedback (RLHF)~\citep{christiano2017deep}, which involves training a reward model based on human feedback. 
This reward model is then utilized within a Proximal Policy Optimization (PPO) framework~\citep{schulman2017proximal} to fine-tune LLMs, ensuring their outputs align more closely with human preferences. 
However, this complex pipeline often suffers from instability and poor sample efficiency~\citep{zhu2023fine}
To address these issues, several enhanced offline approaches have been proposed, such as SLiC~\citep{zhao2022calibrating}, SLiC-HF~\citep{zhao2023slic} and so on. 
Direct preference optimization (DPO)~\citep{rafailov2024direct} directly optimizes the language model by directly training from the policy differences between chosen and rejected pairs.
More simple and direct alignment variants are proposed afterwards~\citep{meng2024simpo, ethayarajh2024kto}. 
Specifically, curry-DPO~\citep{pattnaik2024currydpoenhancingalignmentusing} suggests that using diverse featured data can significantly impact alignment performance. 
By integrating curriculum learning with DPO, it effectively enhances performance.

\textbf{Preference Data Optimization.}
The preference data often encompasses a \textit{prompt}, a \textit{chosen response}, and a \textit{rejected response}~\cite{tan2024large}.
As underscored in~\citep{morimura2024filtered}, typical offline alignment methods are highly sensitive to the quality and coverage of preference data~\citep{kalai2024calibratedlanguagemodelshallucinate}. 
Moreover, selecting a dataset with the right balance of size, complexity, quality, and diversity can significantly impact alignment effectiveness~\citep{liu2023makes, zhou2024lima, zhao2023preliminary, wei2023instructiongpt, song2024scaling, xiong2024deliberate}.
Hence, more representative user queries and corresponding optimized preference pairs might be more helpful and lead to less propagated problems~\citep{song2024scalingdatadiversityfinetuning}. 
In this way, many previous work utilize a reward model to rank and select preference pairs~\citep{cui2024ultrafeedback, zhu20237b, liu2023makes, morimura2024filtered}
Notably, \citet{liu2024statisticalrejectionsamplingimproves} statistically proves utilizing rejection sampling to select preference pairs can greatly benefit to a better alignment performance. 
\citet{khaki2024rsdpohybridrejectionsampling, touvron2023llama} systematically combine rejection sampling when constructing pairs, which shows remarkable improvements of both performance and data efficiency.

\section{Conclusions}

In this work, we expose the potential imbalance between knowledge breadth and depth learning within preference learning datasets, and we conduct preliminary experiments using a simple balancing method to validate this observation.
Furthermore, we propose a novel Balance Preference Optimization (\textsc{BPO}) method, which dynamically enhances the knowledge depth learning resources for each sample. \textsc{BPO} measures the knowledge depth of each sample from a unique model optimization perspective and dynamically selects varying numbers of response pairs for each preference sample.
Our experimental results on several alignment benchmarks demonstrate that \textsc{BPO} outperforms other preference optimization methods while maintaining high training efficiency. Additionally, we provide an in-depth analysis of \textsc{BPO}, offering insights and guidelines on balancing preference learning datasets for future research.


\section*{Limitations and Future Work}
Despite the promising results of our hierarchical clustering-based framework, several limitations must be acknowledged. 
First, learning directly from the model's generated responses leads to limited or marginal performance improvement, indicating inherent constraints in self-optimization.
To achieve high-quality data curation, we rely on the GPT-4 model to score responses to construct response pairs, which introduces a certain level of dependence on the performance and potential biases of GPT-4. 

For future work, it is crucial to develop statistical or empirical methods to evaluate whether a user query would be beneficial to the entire alignment pipeline. 
Additionally, refining the generation and construction of preference pairs is essential, as it fundamentally shapes the alignment learning space. 
Further exploration into these areas could yield significant advancements in the alignment capabilities of LLMs.

\bibliography{custom}

\begin{thebibliography}{51}
\providecommand{\natexlab}[1]{#1}

\bibitem[{Albalak et~al.(2024)Albalak, Elazar, Xie, Longpre, Lambert, Wang, Muennighoff, Hou, Pan, Jeong et~al.}]{albalak2024survey}
Alon Albalak, Yanai Elazar, Sang~Michael Xie, Shayne Longpre, Nathan Lambert, Xinyi Wang, Niklas Muennighoff, Bairu Hou, Liangming Pan, Haewon Jeong, et~al. 2024.
\newblock A survey on data selection for language models.
\newblock \emph{arXiv preprint arXiv:2402.16827}.

\bibitem[{Bai et~al.(2022)Bai, Jones, Ndousse, Askell, Chen, DasSarma, Drain, Fort, Ganguli, Henighan et~al.}]{bai2022training}
Yuntao Bai, Andy Jones, Kamal Ndousse, Amanda Askell, Anna Chen, Nova DasSarma, Dawn Drain, Stanislav Fort, Deep Ganguli, Tom Henighan, et~al. 2022.
\newblock Training a helpful and harmless assistant with reinforcement learning from human feedback.
\newblock \emph{arXiv preprint arXiv:2204.05862}.

\bibitem[{Bai et~al.(2024)Bai, Ying, Cao, Lv, He, Wang, Yu, Zeng, Xiao, Lyu et~al.}]{bai2024benchmarking}
Yushi Bai, Jiahao Ying, Yixin Cao, Xin Lv, Yuze He, Xiaozhi Wang, Jifan Yu, Kaisheng Zeng, Yijia Xiao, Haozhe Lyu, et~al. 2024.
\newblock Benchmarking foundation models with language-model-as-an-examiner.
\newblock \emph{Advances in Neural Information Processing Systems}, 36.

\bibitem[{Christiano et~al.(2017)Christiano, Leike, Brown, Martic, Legg, and Amodei}]{christiano2017deep}
Paul~F Christiano, Jan Leike, Tom Brown, Miljan Martic, Shane Legg, and Dario Amodei. 2017.
\newblock Deep reinforcement learning from human preferences.
\newblock \emph{Advances in neural information processing systems}, 30.

\bibitem[{Cui et~al.(2024)Cui, Yuan, Ding, Yao, He, Zhu, Ni, Xie, Xie, Lin et~al.}]{cui2024ultrafeedback}
Ganqu Cui, Lifan Yuan, Ning Ding, Guanming Yao, Bingxiang He, Wei Zhu, Yuan Ni, Guotong Xie, Ruobing Xie, Yankai Lin, et~al. 2024.
\newblock Ultrafeedback: Boosting language models with scaled ai feedback.
\newblock In \emph{Forty-first International Conference on Machine Learning}.

\bibitem[{Cui et~al.(2023)Cui, Yuan, Ding, Yao, Zhu, Ni, Xie, Liu, and Sun}]{cui2023ultrafeedback}
Ganqu Cui, Lifan Yuan, Ning Ding, Guanming Yao, Wei Zhu, Yuan Ni, Guotong Xie, Zhiyuan Liu, and Maosong Sun. 2023.
\newblock Ultrafeedback: Boosting language models with high-quality feedback.

\bibitem[{Dai et~al.(2023)Dai, Pan, Sun, Ji, Xu, Liu, Wang, and Yang}]{dai2023safe}
Josef Dai, Xuehai Pan, Ruiyang Sun, Jiaming Ji, Xinbo Xu, Mickel Liu, Yizhou Wang, and Yaodong Yang. 2023.
\newblock Safe rlhf: Safe reinforcement learning from human feedback.
\newblock \emph{arXiv preprint arXiv:2310.12773}.

\bibitem[{Dubey et~al.(2024)Dubey, Jauhri, Pandey, Kadian, Al-Dahle, Letman, Mathur, Schelten, Yang, Fan et~al.}]{dubey2024llama}
Abhimanyu Dubey, Abhinav Jauhri, Abhinav Pandey, Abhishek Kadian, Ahmad Al-Dahle, Aiesha Letman, Akhil Mathur, Alan Schelten, Amy Yang, Angela Fan, et~al. 2024.
\newblock The llama 3 herd of models.
\newblock \emph{arXiv preprint arXiv:2407.21783}.

\bibitem[{Dubois et~al.(2024)Dubois, Galambosi, Liang, and Hashimoto}]{dubois2024length}
Yann Dubois, Bal{\'a}zs Galambosi, Percy Liang, and Tatsunori~B Hashimoto. 2024.
\newblock Length-controlled alpacaeval: A simple way to debias automatic evaluators.
\newblock \emph{arXiv preprint arXiv:2404.04475}.

\bibitem[{Ethayarajh et~al.(2024)Ethayarajh, Xu, Muennighoff, Jurafsky, and Kiela}]{ethayarajh2024kto}
Kawin Ethayarajh, Winnie Xu, Niklas Muennighoff, Dan Jurafsky, and Douwe Kiela. 2024.
\newblock Kto: Model alignment as prospect theoretic optimization.
\newblock \emph{arXiv preprint arXiv:2402.01306}.

\bibitem[{Gao et~al.(2023)Gao, Ruan, Sun, Yin, Yang, and Wan}]{gao2023human}
Mingqi Gao, Jie Ruan, Renliang Sun, Xunjian Yin, Shiping Yang, and Xiaojun Wan. 2023.
\newblock Human-like summarization evaluation with chatgpt.
\newblock \emph{arXiv preprint arXiv:2304.02554}.

\bibitem[{Han et~al.(2023)Han, Simig, Mihaylov, Tsvetkov, Celikyilmaz, and Wang}]{han2023understanding}
Xiaochuang Han, Daniel Simig, Todor Mihaylov, Yulia Tsvetkov, Asli Celikyilmaz, and Tianlu Wang. 2023.
\newblock Understanding in-context learning via supportive pretraining data.
\newblock In \emph{Proceedings of the 61st Annual Meeting of the Association for Computational Linguistics (Volume 1: Long Papers)}, pages 12660--12673.

\bibitem[{Hu et~al.(2021)Hu, Shen, Wallis, Allen-Zhu, Li, Wang, Wang, and Chen}]{hu2021lora}
Edward~J Hu, Yelong Shen, Phillip Wallis, Zeyuan Allen-Zhu, Yuanzhi Li, Shean Wang, Lu~Wang, and Weizhu Chen. 2021.
\newblock Lora: Low-rank adaptation of large language models.
\newblock \emph{arXiv preprint arXiv:2106.09685}.

\bibitem[{Johnson(1984)}]{johnson1984extensions}
William~B Johnson. 1984.
\newblock Extensions of lipshitz mapping into hilbert space.
\newblock In \emph{Conference modern analysis and probability, 1984}, pages 189--206.

\bibitem[{Kalai and Vempala(2024)}]{kalai2024calibratedlanguagemodelshallucinate}
Adam~Tauman Kalai and Santosh~S. Vempala. 2024.
\newblock \href {https://arxiv.org/abs/2311.14648} {Calibrated language models must hallucinate}.
\newblock \emph{Preprint}, arXiv:2311.14648.

\bibitem[{Khaki et~al.(2024{\natexlab{a}})Khaki, Li, Ma, Yang, and Ramachandra}]{khaki2024rs}
Saeed Khaki, JinJin Li, Lan Ma, Liu Yang, and Prathap Ramachandra. 2024{\natexlab{a}}.
\newblock Rs-dpo: A hybrid rejection sampling and direct preference optimization method for alignment of large language models.
\newblock \emph{arXiv preprint arXiv:2402.10038}.

\bibitem[{Khaki et~al.(2024{\natexlab{b}})Khaki, Li, Ma, Yang, and Ramachandra}]{khaki2024rsdpohybridrejectionsampling}
Saeed Khaki, JinJin Li, Lan Ma, Liu Yang, and Prathap Ramachandra. 2024{\natexlab{b}}.
\newblock \href {https://arxiv.org/abs/2402.10038} {Rs-dpo: A hybrid rejection sampling and direct preference optimization method for alignment of large language models}.
\newblock \emph{Preprint}, arXiv:2402.10038.

\bibitem[{Kingma(2014)}]{kingma2014adam}
Diederik~P Kingma. 2014.
\newblock Adam: A method for stochastic optimization.
\newblock \emph{arXiv preprint arXiv:1412.6980}.

\bibitem[{Li et~al.(2024{\natexlab{a}})Li, Yang, Tan, Baik, Yun, Lee, Chacko, Hou, Duong-Tran, Ding et~al.}]{li2024dalk}
Dawei Li, Shu Yang, Zhen Tan, Jae~Young Baik, Sunkwon Yun, Joseph Lee, Aaron Chacko, Bojian Hou, Duy Duong-Tran, Ying Ding, et~al. 2024{\natexlab{a}}.
\newblock Dalk: Dynamic co-augmentation of llms and kg to answer alzheimer's disease questions with scientific literature.
\newblock \emph{arXiv preprint arXiv:2405.04819}.

\bibitem[{Li et~al.(2024{\natexlab{b}})Li, Chiang, Frick, Dunlap, Wu, Zhu, Gonzalez, and Stoica}]{li2024crowdsourced}
Tianle Li, Wei-Lin Chiang, Evan Frick, Lisa Dunlap, Tianhao Wu, Banghua Zhu, Joseph~E Gonzalez, and Ion Stoica. 2024{\natexlab{b}}.
\newblock From crowdsourced data to high-quality benchmarks: Arena-hard and benchbuilder pipeline.
\newblock \emph{arXiv preprint arXiv:2406.11939}.

\bibitem[{Li et~al.(2023)Li, Zhang, Dubois, Taori, Gulrajani, Guestrin, Liang, and Hashimoto}]{li2023alpacaeval}
Xuechen Li, Tianyi Zhang, Yann Dubois, Rohan Taori, Ishaan Gulrajani, Carlos Guestrin, Percy Liang, and Tatsunori~B Hashimoto. 2023.
\newblock Alpacaeval: An automatic evaluator of instruction-following models.

\bibitem[{Liu et~al.(2024)Liu, Zhao, Joshi, Khalman, Saleh, Liu, and Liu}]{liu2024statisticalrejectionsamplingimproves}
Tianqi Liu, Yao Zhao, Rishabh Joshi, Misha Khalman, Mohammad Saleh, Peter~J. Liu, and Jialu Liu. 2024.
\newblock \href {https://arxiv.org/abs/2309.06657} {Statistical rejection sampling improves preference optimization}.
\newblock \emph{Preprint}, arXiv:2309.06657.

\bibitem[{Liu et~al.(2023)Liu, Zeng, He, Jiang, and He}]{liu2023makes}
Wei Liu, Weihao Zeng, Keqing He, Yong Jiang, and Junxian He. 2023.
\newblock What makes good data for alignment? a comprehensive study of automatic data selection in instruction tuning.
\newblock \emph{arXiv preprint arXiv:2312.15685}.

\bibitem[{Meng et~al.(2024)Meng, Xia, and Chen}]{meng2024simpo}
Yu~Meng, Mengzhou Xia, and Danqi Chen. 2024.
\newblock Simpo: Simple preference optimization with a reference-free reward.
\newblock \emph{arXiv preprint arXiv:2405.14734}.

\bibitem[{Morimura et~al.(2024)Morimura, Sakamoto, Jinnai, Abe, and Air}]{morimura2024filtered}
Tetsuro Morimura, Mitsuki Sakamoto, Yuu Jinnai, Kenshi Abe, and Kaito Air. 2024.
\newblock Filtered direct preference optimization.
\newblock \emph{arXiv preprint arXiv:2404.13846}.

\bibitem[{Pattnaik et~al.(2024{\natexlab{a}})Pattnaik, Maheshwary, Ogueji, Yadav, and Madhusudhan}]{pattnaik2024curry}
Pulkit Pattnaik, Rishabh Maheshwary, Kelechi Ogueji, Vikas Yadav, and Sathwik~Tejaswi Madhusudhan. 2024{\natexlab{a}}.
\newblock Curry-dpo: Enhancing alignment using curriculum learning \& ranked preferences.
\newblock \emph{arXiv preprint arXiv:2403.07230}.

\bibitem[{Pattnaik et~al.(2024{\natexlab{b}})Pattnaik, Maheshwary, Ogueji, Yadav, and Madhusudhan}]{Pattnaik2024CurryDPOEA}
Pulkit Pattnaik, Rishabh Maheshwary, Kelechi Ogueji, Vikas Yadav, and Sathwik~Tejaswi Madhusudhan. 2024{\natexlab{b}}.
\newblock \href {https://api.semanticscholar.org/CorpusID:268364003} {Curry-dpo: Enhancing alignment using curriculum learning \& ranked preferences}.
\newblock \emph{ArXiv}, abs/2403.07230.

\bibitem[{Pattnaik et~al.(2024{\natexlab{c}})Pattnaik, Maheshwary, Ogueji, Yadav, and Madhusudhan}]{pattnaik2024currydpoenhancingalignmentusing}
Pulkit Pattnaik, Rishabh Maheshwary, Kelechi Ogueji, Vikas Yadav, and Sathwik~Tejaswi Madhusudhan. 2024{\natexlab{c}}.
\newblock \href {https://arxiv.org/abs/2403.07230} {Curry-dpo: Enhancing alignment using curriculum learning \& ranked preferences}.
\newblock \emph{Preprint}, arXiv:2403.07230.

\bibitem[{Petukhova et~al.(2024)Petukhova, Matos-Carvalho, and Fachada}]{petukhova2024text}
Alina Petukhova, Joao~P Matos-Carvalho, and Nuno Fachada. 2024.
\newblock Text clustering with llm embeddings.
\newblock \emph{arXiv preprint arXiv:2403.15112}.

\bibitem[{Pruthi et~al.(2020)Pruthi, Liu, Kale, and Sundararajan}]{pruthi2020estimating}
Garima Pruthi, Frederick Liu, Satyen Kale, and Mukund Sundararajan. 2020.
\newblock Estimating training data influence by tracing gradient descent.
\newblock \emph{Advances in Neural Information Processing Systems}, 33:19920--19930.

\bibitem[{Rafailov et~al.(2024)Rafailov, Sharma, Mitchell, Manning, Ermon, and Finn}]{rafailov2024direct}
Rafael Rafailov, Archit Sharma, Eric Mitchell, Christopher~D Manning, Stefano Ermon, and Chelsea Finn. 2024.
\newblock Direct preference optimization: Your language model is secretly a reward model.
\newblock \emph{Advances in Neural Information Processing Systems}, 36.

\bibitem[{Saeidi et~al.(2024)Saeidi, Verma, and Baral}]{saeidi2024insights}
Amir Saeidi, Shivanshu Verma, and Chitta Baral. 2024.
\newblock Insights into alignment: Evaluating dpo and its variants across multiple tasks.
\newblock \emph{arXiv preprint arXiv:2404.14723}.

\bibitem[{Schulman et~al.(2017)Schulman, Wolski, Dhariwal, Radford, and Klimov}]{schulman2017proximal}
John Schulman, Filip Wolski, Prafulla Dhariwal, Alec Radford, and Oleg Klimov. 2017.
\newblock Proximal policy optimization algorithms.
\newblock \emph{arXiv preprint arXiv:1707.06347}.

\bibitem[{Shen et~al.(2024)Shen, Chen, Backes, Shen, and Zhang}]{shen2024donowcharacterizingevaluating}
Xinyue Shen, Zeyuan Chen, Michael Backes, Yun Shen, and Yang Zhang. 2024.
\newblock \href {https://arxiv.org/abs/2308.03825} {"do anything now": Characterizing and evaluating in-the-wild jailbreak prompts on large language models}.
\newblock \emph{Preprint}, arXiv:2308.03825.

\bibitem[{Song et~al.(2024{\natexlab{a}})Song, Yu, Lang, Yu, Huang, Wang, and Li}]{song2024scaling}
Feifan Song, Bowen Yu, Hao Lang, Haiyang Yu, Fei Huang, Houfeng Wang, and Yongbin Li. 2024{\natexlab{a}}.
\newblock Scaling data diversity for fine-tuning language models in human alignment.
\newblock \emph{arXiv preprint arXiv:2403.11124}.

\bibitem[{Song et~al.(2024{\natexlab{b}})Song, Yu, Lang, Yu, Huang, Wang, and Li}]{song2024scalingdatadiversityfinetuning}
Feifan Song, Bowen Yu, Hao Lang, Haiyang Yu, Fei Huang, Houfeng Wang, and Yongbin Li. 2024{\natexlab{b}}.
\newblock \href {https://arxiv.org/abs/2403.11124} {Scaling data diversity for fine-tuning language models in human alignment}.
\newblock \emph{Preprint}, arXiv:2403.11124.

\bibitem[{Tan et~al.(2024)Tan, Beigi, Wang, Guo, Bhattacharjee, Jiang, Karami, Li, Cheng, and Liu}]{tan2024large}
Zhen Tan, Alimohammad Beigi, Song Wang, Ruocheng Guo, Amrita Bhattacharjee, Bohan Jiang, Mansooreh Karami, Jundong Li, Lu~Cheng, and Huan Liu. 2024.
\newblock Large language models for data annotation: A survey.
\newblock \emph{arXiv preprint arXiv:2402.13446}.

\bibitem[{Touvron et~al.(2023)Touvron, Martin, Stone, Albert, Almahairi, Babaei, Bashlykov, Batra, Bhargava, Bhosale et~al.}]{touvron2023llama}
Hugo Touvron, Louis Martin, Kevin Stone, Peter Albert, Amjad Almahairi, Yasmine Babaei, Nikolay Bashlykov, Soumya Batra, Prajjwal Bhargava, Shruti Bhosale, et~al. 2023.
\newblock Llama 2: Open foundation and fine-tuned chat models.
\newblock \emph{arXiv preprint arXiv:2307.09288}.

\bibitem[{Wang et~al.(2024)Wang, Li, and Lu}]{wang2024self}
Tianduo Wang, Shichen Li, and Wei Lu. 2024.
\newblock Self-training with direct preference optimization improves chain-of-thought reasoning.
\newblock \emph{arXiv preprint arXiv:2407.18248}.

\bibitem[{Wei et~al.(2023)Wei, Jiang, Huang, and Sun}]{wei2023instructiongpt}
Lai Wei, Zihao Jiang, Weiran Huang, and Lichao Sun. 2023.
\newblock Instructiongpt-4: A 200-instruction paradigm for fine-tuning minigpt-4.
\newblock \emph{arXiv preprint arXiv:2308.12067}.

\bibitem[{Xia et~al.(2024)Xia, Malladi, Gururangan, Arora, and Chen}]{xia2024less}
Mengzhou Xia, Sadhika Malladi, Suchin Gururangan, Sanjeev Arora, and Danqi Chen. 2024.
\newblock Less: Selecting influential data for targeted instruction tuning.
\newblock \emph{arXiv preprint arXiv:2402.04333}.

\bibitem[{Xiong et~al.(2024)Xiong, Payani, Yang, and Fekri}]{xiong2024deliberate}
Siheng Xiong, Ali Payani, Yuan Yang, and Faramarz Fekri. 2024.
\newblock Deliberate reasoning for llms as structure-aware planning with accurate world model.
\newblock \emph{arXiv preprint arXiv:2410.03136}.

\bibitem[{Zhang et~al.(2024)Zhang, Wu, Li, Yang, Zhao, Jiang, and Tan}]{zhang2024balancing}
Hengyuan Zhang, Yanru Wu, Dawei Li, Zacc Yang, Rui Zhao, Yong Jiang, and Fei Tan. 2024.
\newblock Balancing speciality and versatility: a coarse to fine framework for supervised fine-tuning large language model.
\newblock \emph{arXiv preprint arXiv:2404.10306}.

\bibitem[{Zhao et~al.(2023{\natexlab{a}})Zhao, Joshi, Liu, Khalman, Saleh, and Liu}]{zhao2023slic}
Yao Zhao, Rishabh Joshi, Tianqi Liu, Misha Khalman, Mohammad Saleh, and Peter~J Liu. 2023{\natexlab{a}}.
\newblock Slic-hf: Sequence likelihood calibration with human feedback.
\newblock \emph{arXiv preprint arXiv:2305.10425}.

\bibitem[{Zhao et~al.(2022)Zhao, Khalman, Joshi, Narayan, Saleh, and Liu}]{zhao2022calibrating}
Yao Zhao, Mikhail Khalman, Rishabh Joshi, Shashi Narayan, Mohammad Saleh, and Peter~J Liu. 2022.
\newblock Calibrating sequence likelihood improves conditional language generation.
\newblock In \emph{The eleventh international conference on learning representations}.

\bibitem[{Zhao et~al.(2023{\natexlab{b}})Zhao, Yu, Hui, Yu, Huang, Li, and Zhang}]{zhao2023preliminary}
Yingxiu Zhao, Bowen Yu, Binyuan Hui, Haiyang Yu, Fei Huang, Yongbin Li, and Nevin~L Zhang. 2023{\natexlab{b}}.
\newblock A preliminary study of the intrinsic relationship between complexity and alignment.
\newblock \emph{arXiv preprint arXiv:2308.05696}.

\bibitem[{Zheng et~al.(2023{\natexlab{a}})Zheng, Chiang, Sheng, Zhuang, Wu, Zhuang, Lin, Li, Li, Xing et~al.}]{zheng2023judging}
Lianmin Zheng, Wei-Lin Chiang, Ying Sheng, Siyuan Zhuang, Zhanghao Wu, Yonghao Zhuang, Zi~Lin, Zhuohan Li, Dacheng Li, Eric Xing, et~al. 2023{\natexlab{a}}.
\newblock Judging llm-as-a-judge with mt-bench and chatbot arena.
\newblock \emph{Advances in Neural Information Processing Systems}, 36:46595--46623.

\bibitem[{Zheng et~al.(2023{\natexlab{b}})Zheng, Chiang, Sheng, Zhuang, Wu, Zhuang, Lin, Li, Li, Xing, Zhang, Gonzalez, and Stoica}]{zheng2023judgingllmasajudgemtbenchchatbot}
Lianmin Zheng, Wei-Lin Chiang, Ying Sheng, Siyuan Zhuang, Zhanghao Wu, Yonghao Zhuang, Zi~Lin, Zhuohan Li, Dacheng Li, Eric~P. Xing, Hao Zhang, Joseph~E. Gonzalez, and Ion Stoica. 2023{\natexlab{b}}.
\newblock \href {https://arxiv.org/abs/2306.05685} {Judging llm-as-a-judge with mt-bench and chatbot arena}.
\newblock \emph{Preprint}, arXiv:2306.05685.

\bibitem[{Zhou et~al.(2024)Zhou, Liu, Xu, Iyer, Sun, Mao, Ma, Efrat, Yu, Yu et~al.}]{zhou2024lima}
Chunting Zhou, Pengfei Liu, Puxin Xu, Srinivasan Iyer, Jiao Sun, Yuning Mao, Xuezhe Ma, Avia Efrat, Ping Yu, Lili Yu, et~al. 2024.
\newblock Lima: Less is more for alignment.
\newblock \emph{Advances in Neural Information Processing Systems}, 36.

\bibitem[{Zhu et~al.(2023{\natexlab{a}})Zhu, Frick, Wu, Zhu, and Jiao}]{zhu20237b}
B~Zhu, E~Frick, T~Wu, H~Zhu, and J~Starling Jiao. 2023{\natexlab{a}}.
\newblock 7b: Improving llm helpfulness \& harmlessness with rlaif.

\bibitem[{Zhu et~al.(2023{\natexlab{b}})Zhu, Sharma, Frujeri, Dong, Zhu, Jordan, and Jiao}]{zhu2023fine}
Banghua Zhu, Hiteshi Sharma, Felipe~Vieira Frujeri, Shi Dong, Chenguang Zhu, Michael~I Jordan, and Jiantao Jiao. 2023{\natexlab{b}}.
\newblock Fine-tuning language models with advantage-induced policy alignment.
\newblock \emph{arXiv preprint arXiv:2306.02231}.

\end{thebibliography}

\appendix
\onecolumn

\section{Additional Results for Preliminary Experiment}
\label{app:preliminary}

\begin{table*}[h]
\centering
\setlength\tabcolsep{4pt}
\fontsize{8.5}{8}\selectfont 
\begin{tabular}{cccccc}
\toprule[2pt]
\addlinespace[2pt]
\multirow{2}{*}{\textbf{Scaling Ratio}} & \multicolumn{2}{c}{\textbf{safeRLHF}} & & \multicolumn{2}{c}{\textbf{HH-RLHF}} \\ \addlinespace[2pt] \cline{2-3}  \cline{5-6} \addlinespace[2pt]
& \textbf{MT-Bench (Avg Score)}  & \textbf{AlpacaEval (win \%)} & & \textbf{MT-Bench (Avg Score)} & \textbf{AlpacaEval (win \%)} \\ \addlinespace[2pt] \hline \addlinespace[2pt]
Fullset        & 4.90   & 36.84\% & & 4.71  & 18.14\% \\ \addlinespace[2pt] \hdashline[1pt/1pt] \addlinespace[2pt]
1\%            & 5.07   & 40.41\% & & 4.17  & 17.14\% \\ 
1\% Simple Balance         & 5.17   & 41.22\% & & 4.29  & 21.41\% \\ 
\addlinespace[2pt] \hdashline[1pt/1pt] \addlinespace[2pt]

5\%            & 4.99   & 42.99\%  & & 4.22  & 20.87\% \\
5\% Simple Balance          & 5.03   & 42.43\%  & & 4.34  & 24.61\% \\ 
\addlinespace[2pt] \hdashline[1pt/1pt] \addlinespace[2pt]

10\%           & 5.12   & 45.75\% & & 4.11  & 20.12\% \\  
10\% Simple Balance         & 5.19  & 46.18\%  & & 4.42  & 25.34\% \\ 
\addlinespace[2pt] \hdashline[1pt/1pt] \addlinespace[2pt]

20\%           & 5.05   & 42.70\% & & 3.99  & 16.77\% \\
20\% Simple Balance         & 5.10   & 45.47\% & & 4.36  & 22.30\% \\

\bottomrule[2pt]

\end{tabular}

\caption{
Performance of the embedding-based clustering method across different prompt selection percentages.
The performance is evaluated on MT-Bench with the average scores of round 1 and round 2, and the win rate on AlpacaEval against text-davinci-003.
The results show that almost only 1\% to 10\% of the prompts can lead to promising outcomes.
For all experiments, the base model is $\text{Llama-3}_{\text{8B}}$, which undergoes preliminary SFT on the same seed data, followed by DPO. 
}
\label{tab: less prompt}
\end{table*}

\section{Preliminary Supervised Fine-Tuning}
\label{preSFT}
The preliminary step involves fine-tuning the unsupervised base model \(\pi_0\) using a small, high-quality annotated seed data \( \mathcal{D}_{seed} = \{(x_1, y_1), \dots, (x_m, y_m)\} \), where \( x_i \) is the input text and \( y_i \) is the corresponding target output.
Additionally, we use LoRA, which freezes the pre-trained model weights and injects trainable rank decomposition matrices into each layer, to reduce parameters that need to be trained \citep{hu2021lora}.
This supervised fine-tuning (SFT) step effectively refines the base model, enhancing the subsequent DPO performance \citep{wang2024self,saeidi2024insights}. 
The fine-tuning objective is to minimize the supervised loss:

\begin{equation}
\small
\mathcal{L}_{\text{SFT}}(\theta) = \mathbb{E}_{(x_i, y_i) \sim \mathcal{D}_\text{seed}} \left[ - \log \pi_\theta(y_i \mid x_i) \right]
\label{sft}
\end{equation}

where \( \theta \) represents the model parameters.

\section{Random Project for Gradients}
\label{app:random project}
Following \citet{xia2024less}, for a given data sample \( z \) and model \( \theta \), we compute a \( d \)-dimensional projection of the LoRA gradient \(\widehat{\nabla} \ell (z; \theta_t) = \Pi^\top \nabla \ell (z; \theta_t)\), where each entry of \( \Pi \in \mathbb{R}^{P \times d} \) is drawn from a Rademacher distribution \citep{johnson1984extensions} (i.e., \( \Pi_{ij} \sim \mathcal{U}(\{-1, 1\}) \)). 
In total, we compute the projected gradient features for each data sample \( z \) as \( \widetilde{\Gamma}(z, \cdot) = \Pi^\top \Gamma(z, \theta_t) \).

\section{Additional Results for Dynamic Knowledge Depth Augmentation}
\begin{table}[h]
\small
\centering
\begin{tabular}{cccc}
\toprule[2pt]
\textbf{Number of Clusters} & \textbf{\(\eta\%\)} & \textbf{MT-Bench} & \textbf{AlpacaEval} \\  \addlinespace[2pt] 
 \hline \addlinespace[2pt] 
\multirow{3}{*}{50} & 1\% & 5.19 & 26.46\% \\ 
 & 5\% & 5.32 & 33.79\% \\ 
& 10\% & 5.48 & 40.25\% \\  \hline
\multirow{3}{*}{100}    
& 1\%       & 4.30  & 25.59\% \\ 
 & 5\%      & 4.67  & 32.17\% \\ 
& 10\%      & 5.29 & 34.16\% \\ 
 \bottomrule[2pt]
\end{tabular}
\caption{
Influences of different \(\eta \%\) to dynamic knowledge depth augmentation stage in our method, \textsc{BPO}. Clustering method is \textit{K}-means, and the number of clusters is 50 and 100. 
Experiments were conducted on Llama-3-8B, HH-RLHF dataset.
}
\label{ab:clustering about kpa}
\end{table}

\section{Length Differences}
\label{ab:length}
For generated responses, we consider their token length to  measure the required knowledge depth.
For each prompt \( x_i \), let \( \{ y_{i1}, y_{i2}, \dots, y_{ik} \} \) denote the generated candidate responses. 
We compute the token length \( l_{ij} \) of each response \( y_{ij} \). 
The sample variance of the lengths for prompt \( x_i \) is calculated as:
\begin{equation}
\sigma^{2}_{l_i} = \frac{1}{k-1} \sum_{j=1}^{k} (l_{ij} - \bar{l}_i )^{2}
\end{equation}
where \( \bar{l}_i = \frac{1}{k} \sum_{j=1}^{k} l_{ij} \) is the mean length of the responses for prompt \( x_i \). 
This variance \( \sigma^{2}_{l_i} \) reflects the diversity in response lengths, indicating the consistency level of understanding.
These variances are normalized to obtain the weight \(w_i^{l}\) for each prompt:
\begin{equation} 
w_i^{{l}} = \frac{\sigma^{2}_{l_i}}{\sum_{i=1}^{n^{'}} \sigma^{2}_{l_i}} 
\end{equation}
where \( n^{'} \) is the number of prompts in \( \mathcal{X}_{rep} \).
Based on this, we dynamically allocate the depth \( k_i \) for each prompt \( x_i \), ensuring that prompts with greater variability receive more training samples. 
The depth \( k_i \) is determined as: \(k_i = \left\lceil w_i \times N \right\rceil\),
where \( N \) is the total number of responses to be allocated across all prompts, and \( \lceil \cdot \rceil \) denotes the ceiling function.

\section{Semantic Similarity Difference}
\label{ab:semantic}

We also leverage semantic similarity to measure the required knowledge depth.
For each prompt \( x_i \), let \( \{ y_{i1}, y_{i2}, \dots, y_{ik} \} \) denote the set of generated candidate responses.
We obtain the embedding vector \( \mathbf{e}_{ij} \) for each response \( y_{ij} \) using a pre-trained embedder \( f \).
We compute the pairwise cosine similarities between all response embeddings and calculate the average cosine similarity \( \bar{s}_i \) for prompt \( x_i \):
\begin{equation}
\bar{s}_i = \frac{2}{k(k-1)} \sum_{1 \leq p < q \leq k} \cos\left( \mathbf{e}_{ip}, \mathbf{e}_{iq} \right)
\end{equation}
where
\begin{equation}
\cos\left( \mathbf{e}_{ip}, \mathbf{e}_{iq} \right) = \frac{\mathbf{e}_{ip} \cdot \mathbf{e}_{iq}}{\left\| \mathbf{e}_{ip} \right\| \left\| \mathbf{e}_{iq} \right\|}
\end{equation}
denotes the cosine similarity between embeddings \( \mathbf{e}_{ip} \) and \( \mathbf{e}_{iq} \), and \( \left\| \cdot \right\| \) represents the Euclidean norm.
To determine the dynamic depth, we calculate the weight \( w_i^{s} \) for each prompt, which is inversely proportional to the average cosine similarity:
\begin{equation}
w_i^{s} = \frac{\dfrac{1}{\bar{s}_i}}{\sum_{i=1}^{n'} \dfrac{1}{\bar{s}_i}}
\end{equation}
where \( n' \) is the number of prompts in \( \mathcal{X}_{ep} \).
This inverse relationship ensures that prompts with higher overall response similarity (i.e., less semantic diversity) are assigned a smaller depth.
Based on these weights, we allocate the depth \( k_i \) for each prompt \( x_i \) as:
\begin{equation}
k_i = \left\lceil w_i^{s} \times N \right\rceil
\end{equation}
where \( N \) is the total number of responses to be distributed across all prompts, and \( \lceil \cdot \rceil \) denotes the ceiling function.

\section{Dataset Details}
\label{dataset}

\paragraph{Trainset}
HH-RLHF consists of two subsets, with approximately 170K chosen-rejected pairs related to human and AI assistant dialogues. 
We randomly select 30,000 samples from the helpfulness subset for our experiment. 
SafeRLHF involves 83.4K preference entries and each is designed with two responses, accompanied by two labels for harmlessness and helpfulness.
In our experiment, we select samples where the two labels are the same, yielding approximately 55,000 samples. 
Finally, for each dataset, we randomly select 10\% of the dataset \( \mathcal{D}_{org} \) as the seed data \( \mathcal{D}_{seed} \), using annotated chosen responses as golden reference to perform preliminary SFT. 

\paragraph{Evaluation Set}
Specifically, MT-Bench is a multi-turn benchmark that evaluates the capacity of LLMs to engage in coherent, informative, and engaging conversations. 
We present the average score of each round. 
AlpacaEval, on the other hand, is designed to evaluate the win-lose-tie rate of LLMs by comparing their responses to the baseline model, text-davinci-003.

\section{Training Hyperparameters}
\label{hyper}

The learning rate is set to 2e-5, with a batch size of 32 for 4 epochs. 
In the LoRA setting, we use the AdamW optimizer.
When constructing pairs, we follow \citet{khaki2024rs} to generate responses for 10\% of representative prompts sampled through embedding-based clustering as described in Section~\ref{Dynamic Knowledge Depth Augmentation}. 
We set \textit{k} = 16 responses per prompt, with a maximum of 512 new tokens, a top-k value of 50, and a temperature of 1. 
Then, we select the top 10\% of pairs with the highest score differences from all constructed pairs, and compute the gradients of the selected samples. 
For gradient projection, we follow \citet{xia2024less}'s settings to randomly project the gradient features to \textit{d} = 8192. 
In DPO training, we set the learning rate to 2e-5, with a batch size of 16 for 4 epochs.

\clearpage
\section{Additional Experiments}
\label{additional_exp}
\appendix

To further demonstrate that BPO is both effective and robust, we conducteded additional experiments on the UltraFeedback dataset~\cite{cui2023ultrafeedback}, anselectedcted AlpacaEval 2.0~\cite{dubois2024length} and Arena-Hard~\cite{li2024crowdsourced} as our evaluation benchmarks.
All experiments were carried out using the Llama-3-8B-Instruct model.

\paragraph{Trainset} 
UltraFeedback contains AI-generated feedback annotations for a wide variety of prompts, facilitating preference modeling.
For our experiments, we randomly select 30,000 samples from UltraFeedback, of which 10\% are used as seed data \( \mathcal{D}_{seed} \), following the setup described in Section~\ref{experimental setup}.
We used the same training hyperparameters as presented in the Appendix~\ref{hyper}.

\paragraph{Evaluation Set}
AlpacaEval 2.0 comprises 805 questions drawn from five different datasets, while Arena-Hard includes 500 well-defined technical problem-solving queries.
We report all scores according to the respective evaluation protocols of the benchmarks.
For AlpacaEval 2.0, we report both the raw win rate and the length-controlled win rate (LC), where the LC metric is designed to be robust against model verbosity.
For Arena-Hard, we report the win rate compared to the baseline model.

\begin{table}[h]
    \centering
    
    \begin{tabular}{lccc}
        \toprule[2pt]
        \textbf{Method} & \multicolumn{2}{c}{\textbf{AlpacaEval 2.0}} & \textbf{Arena Hard} \\
        & LC Win Rate (\%) & Win Rate (\%) & Win Rate (\%) \\
        \midrule
        Base-model & 22.92 & 22.57 & 20.20 \\
        Vanilla DPO & 28.71 & 27.24 & 33.21 \\
        Curry-DPO & 35.46 & 36.65 & 34.71 \\
        RS-DPO & 41.23 & 40.44 & 37.98 \\
        \textbf{BPO (ours)} & \textbf{44.75} & \textbf{43.68} & \textbf{39.83} \\
        \bottomrule[2pt]
    \end{tabular}
    \caption{Performance comparison of different pair selection strategies on the UltraFeedback dataset.}
\label{tab:ultra_results}
\end{table}
As shown in Table~\ref{tab:ultra_results}, our BPO method demonstrates substantial improvements over the base model and other baselines on both the AlpacaEval 2.0 and Arena Hard datasets. These additional experiments further validate the robustness and effectiveness of BPO across diverse datasets and models.

\clearpage
\section{LLM-as-a-Judge's Prompt}
\label{tab: judge_prompt}

We follow the prompt designed in~\citep{zheng2023judgingllmasajudgemtbenchchatbot}.
Prompts used in AlpacaEval are shown as follow: \\

\begin{mdframed}
\begin{verbatim}
[System] 
Please act as an impartial judge and evaluate the quality of the responses 
provided by two AI assistants to the user question displayed below. 
You should choose the assistant that follows the user’s instructions and answers 
the user’s question better. Your evaluation should consider factors 
such as the helpfulness, relevance, accuracy, depth, creativity, and level of 
detail of their responses. Begin your evaluation by comparing 
the two responses and provide a short explanation. Avoid any position biases 
and ensure that the order in which the responses were presented does not 
influence your decision. Do not allow the length of the responses to influence 
your evaluation. Do not favor certain names of the assistants. Be as objective 
as possible. After providing your explanation, output your final verdict by 
strictly following this format: "[[A]]" if assistant A is better, 
"[[B]]" if assistant B is better, and "[[C]]" for a tie.
[User Question]  {}
[The Start of Assistant A’s Answer] {} [The End of Assistant A’s Answer]
[The Start of Assistant B’s Answer] {} [The End of Assistant B’s Answer] 

\end{verbatim}
\end{mdframed} 

\vspace{1em}

Prompts designed to evaluate models' performance in MT-Bench are attached below:  \\
\begin{mdframed}
\begin{verbatim}
[System] Please act as an impartial judge and evaluate the quality 
of the response provided by an AI assistant to the user question displayed below. 
Your evaluation should consider factors such as the helpfulness, relevance, 
accuracy, depth, creativity, and level of detail of the response. 
Begin your evaluation by providing a short explanation. 
Be as objective as possible. After providing your explanation, 
please rate the response on a scale of 1 to 10 by strictly following this format: 
"[[rating]]", for example: "Rating: [[5]]".
[Question] {} 
[The Start of Assistant's Answer] {} 
[The End of Assistant's Answer]

\end{verbatim}
\end{mdframed}

\section{Acknowledgment of AI Assistance in Writing and Revision}
We utilized ChatGPT-4 for revising and enhancing wording of this paper.

\end{document}